\documentclass[runningheads,a4paper]{llncs}

\usepackage{epstopdf}
\usepackage{url}

\urldef{\mailsa}\path|xiang.zhang3@student.unsw.edu.au|
\urldef{\mailsb}\path|{xiaocong.chen, lina.yao}@unsw.edu.au|
\urldef{\mailsc}\path|{chang.ge, manqing.dong}@student.unsw.edu.au|

\usepackage{cite}
\usepackage{amsmath,amssymb,amsfonts}
\usepackage{algorithmic}
\usepackage{graphicx}
\usepackage{textcomp}
\usepackage{fancyhdr}\def\BibTeX{{\rm B\kern-.05em{\sc i\kern-.025em b}\kern-.08em
    T\kern-.1667em\lower.7ex\hbox{E}\kern-.125emX}}
\usepackage{booktabs} 
\usepackage{colortbl}
\usepackage[ruled, vlined, linesnumbered]{algorithm2e}

\usepackage{comment}
\usepackage{mathtools}
\usepackage{multirow}
\usepackage[dvipsnames]{xcolor}
\usepackage{balance}
\usepackage{bm}
\usepackage{caption}
\usepackage{subcaption}
\usepackage{eucal}

\makeatletter
\def\endthebibliography{%
  \def\@noitemerr{\@latex@warning{Empty `thebibliography' environment}}%
  \endlist
}
\makeatother

\begin{document}

\mainmatter  

\title{Deep Neural Network Hyperparameter Optimization with Orthogonal Array Tuning}

\titlerunning{Deep Neural Network Hyperparameter Optimization}


\author{Xiang Zhang, Xiaocong Chen, Lina Yao, Chang Ge, Manqing Dong}
\authorrunning{Xiang Zhang et.al}
\institute{University of New South Wales, Australia\\
\mailsa\\
}

\maketitle

\begin{abstract}
Deep learning algorithms have achieved excellent performance lately in a wide range of fields (e.g., computer version). However, a severe challenge faced by deep learning is the high dependency on hyper-parameters. The algorithm results may fluctuate dramatically under the different configuration of hyper-parameters. Addressing the above issue, this paper presents an efficient Orthogonal Array Tuning Method (OATM)
for deep learning hyper-parameter tuning. We describe the OATM approach in five detailed steps and elaborate on it using two widely used deep neural network structures (Recurrent Neural Networks and Convolutional Neural Networks). The proposed method is compared to the state-of-the-art hyper-parameter tuning methods including manually (e.g., grid search and random search) and automatically (e.g., Bayesian Optimization) ones. The experiment results state that OATM can significantly save the tuning time compared to the state-of-the-art methods while preserving the satisfying performance.
 
\end{abstract} 

\section{Introduction}
Deep learning has been recently attracting much attention in both academia and industry, due to its excellent performance on various research areas such as computer vision, speech recognition, natural language processing, and brain-computer interface \cite{zhang2017braintyping}. Nevertheless, deep learning faces an important challenge that the performance of the algorithm highly depends on the selection of hyper-parameters. Compared with traditional machine learning algorithms, deep learning requires hyper-parameter tuning more urgently because deep neural networks: 1) have more hyper-parameters to be tunned; 2) have higher dependency on the configuration of hyper-parameters. \cite{zhang2017intent} reports the deep learning classification accuracy dramatically fluctuates from 32.2\% to 92.6\% due to the different selection of hyper-parameters. Therefore, an effective and efficient hyper-parameter tuning method is necessary.

However, most of the existing hyper-parameter tuning methods have some drawbacks. In particular, grid search traverses all the possible combinations of different hyper-parameters, which is a time-consuming and ad-hoc process \cite{NIPS2011_4443}.
Random Search, which is developed based on grid research, set up a grid of hyper-parameter values and selects random combinations to train the algorithm \cite{NIPS2011_4443}. Random search method oversteps some disadvantages of grid search such as time-consuming but meanwhile brings a major disadvantage which cannot converge to the global optimum \cite{Andradottir2015}. The randomly selected hyper-parameter combinations cannot guarantee a steady and competitive result. 
Apart from the manually tuning methods, automated tuning methods being more popular in recent years \cite{NIPS2012_4522}. Bayesian Optimization, a most widely-used automated hyper-parameter tunning approach, attempts to find the global optimum in a minimum number of steps. Nevertheless, the results of Bayesian optimization are sensitive to parameters of the surrogate model and the performance is highly depending on the quality of the learning model \cite{BOlimitation}.

To address the aforementioned issue, we propose the Orthogonal Array Tuning Method (OATM) which can achieve a trade-off of the less tuning time and competitive performance. In detail, the OATM manner is proposed based on Taguchi Approach \cite{taguchi1987system}. The OATM is a highly fractional orthogonal design method that is based on a design matrix and allows the user to consider a selected subset of combinations of multiple factors at multiple levels. Additionally, the OATM is balanced to ensure that all possible values of all hyper-parameters are considered equally. 
Moreover, OATM has been commonly used as an experimental design method in a wide variety of domains like mechanical engineering \cite{nalbant2007application} and
electrical engineering \cite{mahapatra2007optimization}. 
To our best knowledge, our work is {\it the first batch of work} adopting orthogonal array into parameter tuning in deep learning. 

The proposed OATM adopts the orthogonal array to extract the most representative and balanced combinations from the whole set of possible combinations. 
The proposed OATM will be explained in detail in the context of two popular deep learning structures (Section~\ref{sec:results_analysis}). In addition, the OATM is evaluated over three datasets, which demonstrate the universality and adaptability. We notice that source codes performing grid search, random search, and especially Bayesian Optimization on deep learning are hard to online acquire. Thus, we provide the reusable source codes and datasets for reproduction\footnote{The link will be available after the paper is accepted}.

\section{Related Work} 
\label{sec:related_work} 

Currently, there are several widely used tuning methods such as grid search optimization, random search optimization, and Bayesian optimization.
Grid search and random search require all possible values for each parameter whereas Bayesian optimization needs the range for each parameter. \cite{NIPS2015_5872} proposed automated machine learning method based on the efficiency of Bayesian optimization and \cite{NIPS2013_5086} applied multi-task Gaussian processes to Bayesian optimization to enhance the performance of Bayesian Optimization. However, these methods fail in deep learning architectures for which have larger amount of hyper-parameters and the performance highly rely on the configuration.


Apart from the aforementioned methods, the orthogonal array based hyper-parameter tuning already used in a range of research areas such as mechanical engineering and electrical engineering. J.A Ghani et al. \cite{nalbant2007application} applied orthogonal array based approach to optimize the cutting parameters in the end milling.
 S.S. Mahapatra et al. \cite{mahapatra2007optimization} optimized wire electrical discharge machining (WEDM) process parameters by orthogonal array method.

\noindent{\textbf{Summary.}} The traditional methods are not suited for deep learning algorithms while the effectiveness of OATM has been demonstrated in many research topics. Intuitively, we adopt OATM for deep learning hyper-parameter tuning.
To our best knowledge, our work is the first batch of studies in this area.

\section{Orthogonal Array Tuning} 
\label{sec:orthogonal_array_tuning}

In this section, we first provide the background knowledge of orthogonal array, namely, the definition, the compose principles, and the terminology. Then, we report the working procedure of OATM. 

\subsection{Background of Orthogonal Array} 
\label{sub:background_of_orthogonal_array}
\begin{figure}[ht]
\centering
\begin{minipage}[b]{0.5\linewidth}
\centering
\includegraphics[width=\textwidth]{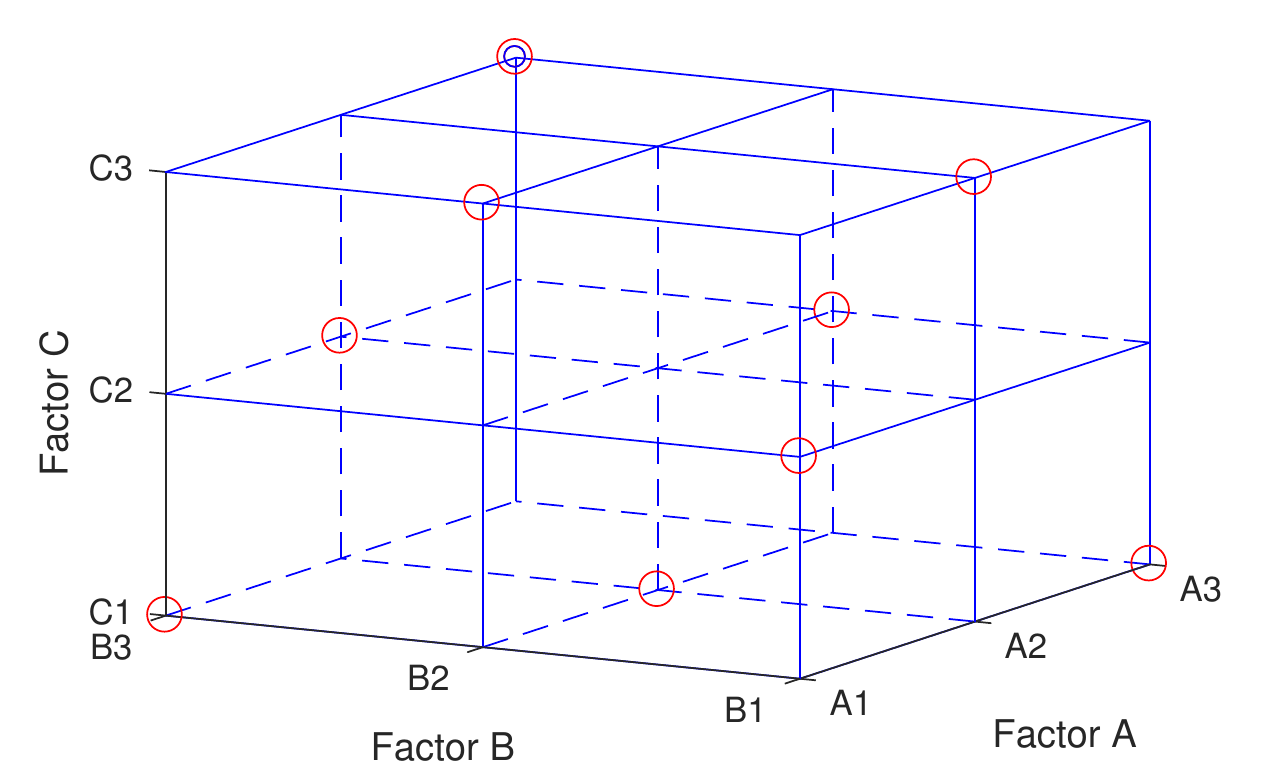}
\caption{Orthogonal Array cube. The red circles are selected combinations by Orthogonal Array.}
\label{fig:cube}
\end{minipage}
\hspace{1mm}
\begin{minipage}[b]{0.4\linewidth}
\centering
\resizebox{\linewidth}{!}{
\begin{tabular}{c|ccc}
\hline
\multirow{2}{*}{\textbf{Row No.}} & \multicolumn{3}{c}{\textbf{Factor No.}} \\ 
 & \textbf{Factor 1} & \textbf{Factor 2} & \textbf{Factor 3} \\ \hline
1 & 1 & 1 & 1 \\
2 & 1 & 2 & 2 \\
3 & 1 & 3 & 3 \\
4 & 2 & 1 & 2 \\
5 & 2 & 2 & 3 \\
6 & 2 & 3 & 1 \\
7 & 3 & 1 & 3 \\
8 & 3 & 2 & 1 \\
9 & 3 & 3 & 2 \\ \hline
\end{tabular}
}
\captionof{table}{Orthogonal Array with 9 rows, 3 factors and each factor has 3 levels}
\label{tab:33fl}
\end{minipage}
\vspace{-3mm}
\end{figure}

An Orthogonal Array is a table/array whose entries come from a fixed finite set of elements (typically, ${1,2,...,n}$), arranged in a specific way that for every selection of two
different columns of the table, all ordered 2-tuples of the elements appear for the same number of times. For example, Table~\ref{tab:33fl} shows an Orthogonal Array whose entries come from a fixed finite set ${1,2,3}$. In the Orthogonal Array, the column is called {\it factor} and each element in the finite set (or the element in each column) is called {\it level}. 

The Orthogonal Array holds two basic composition principles:
\begin{itemize}
\item First, in the same column (factor), different levels have the same appearing times. For example, in the first column of Table~\ref{tab:33fl}, each level (level 1, level 2, and level 3) appears for 3 times. Similarly, in the second and third columns, each level appears for 3 times.
\item Second, in two randomly selected columns (factors), different level combinations are complete and balanced. The number of Orthogonal Array rows is determined by this principle. For example, in the first and second columns of Table~\ref{tab:33fl}, each column has 3 levels and there are totally 9 different ordered combinations: (1,2), (1,3), (1,3), (2,1), (2,2), (2,3), (3,1), (3,2), and (3,3). All the combinations are complete (every combination {\it appears}) and balanced (every combination appears \textit{once}). 
\end{itemize}

The essence of Orthogonal Array is a representative subset of the exhausting full set of the elements. We denote the exhausting combination of all the factors and all the levels(3 factors and 3 levels for the above example) as $\mathcal{S}$. Apparently, $Card(\mathcal{S})=3^3=27$. As shown in Table~\ref{tab:33fl}, the Orthogonal Array only has 9 rows. Let's say $Card(\mathcal{O})=9$, where $\mathcal{O}$ denotes the set of the combinations in the OATM. Easy to know, $\mathcal{O}$ is a representative subset of $\mathcal{S}$, or $\mathcal{O} \subseteq \mathcal{S}$. 
Intuitively, we can draw both sets out in a cube. In Figure~\ref{fig:cube}, $A_1, A_2, A_3$ represent 3 levels of factor $A$, while factors $B, C$ are with the same tokens (factors are supposed to be statistically independent with each other). The total 27 nodes on the surface of the cube denote $\mathcal{S}$ while the 9 circled nodes represent the 9 combinations in $\mathcal{O}$. It's easy to observe in Figure~\ref{fig:cube} that the combinations (circled node) sampled by OATM are uniformly distributed: each edge (totally 27 edges) of the cube has one circled node and each face (totally 9 faces) has three circled nodes.


\subsection{Orthogonal Array Tuning Method } 
\label{sub:orthogonal_array_tuning}
In this section, we propose the Orthogonal Array Tuning Method inspired by the basic principles of orthogonal array. Although deep learning algorithms can achieve good performance in many research areas, tuning the hyper-parameters (e.g., the number of layers, the number of nodes in each layer and the learning rate) is time-consuming and dependent on user's expertise. 

In OATM, the hyper-parameters are regarded as factors and different values of each hyper-parameter are regarded as levels. The procedure is listed as follows.
\begin{itemize}
    \item{\bf Step 1:} Build the F-L (factor-level) table. Determine the number of to-be-tuned factors and the number of levels for each factor. The levels should be determined by experience and literature.
    We further suppose each factor has the same number of levels\footnote{For the sake of simplicity, we consider all the factors with the same number of levels. More advanced knowledge can be found in \cite{taguchi1987system} for more complex situations.}. 
    \item{\bf Step 2:} Construct Orthogonal Array Tuning table. The constructed table should obey the basic composition principles. 
    Here\footnote{\url{https://www.york.ac.uk/depts/maths/tables/taguchi_table.htm}} shows some commonly used tables.
     An alternative way is to use the software. The Orthogonal Array Tuning table can be generated by software such as Weibull++\footnote{\url{http://www.reliasoft.com/Weibull/index.htm}} and SPSS\footnote{\url{https://www.ibm.com/analytics/au/en/technology/spss/}}, more details in this link\footnote{\url{https://www.youtube.com/watch?v=C7PIcOXlWQg}}. The Orthogonal Array Tuning table is marked as $L_M(h^k)$ which has $k$ factors, $h$ levels, and totally M rows.
    \item{\bf Step 3:} Run the experiments with the hyper-parameters determined by the Orthogonal Array Tuning table. 
    \item{\bf Step 4:} Range analysis. This is the key step of OATM. Based on the experiment results in the previous step, range analysis method is employed to analyze the results and figure out the optimal levels and importance of each factor. The importance of a factor is defined by its influence on the results of the experiments. Note that range analysis optimizes each factor and combines the optimal levels together, which means that the optimized hyper-parameter combination is not restricted to the existing Orthogonal Array table.
    \item{\bf Step 5:} Run the experiment with the optimal hyper-parameters setting.
\end{itemize}

The OATM is enabled to optimize the hyper-parameters by utilizing a very small set of highly representative hyper-parameter combinations. The high efficiency can be demonstrated by a simple sample in Figure~\ref{fig:cube}. The OATM only takes 9 combinations (red cycles) which means the hyper-parameters can be optimized by running the experiment for 9 times. In contrast, the grid search requires trying all the 27 combinations (27 nodes in the cube). Therefore, through OATM, we can save about 67\% ($0.67=1-9/27$) work in the tuning procedure.

\section{Experimental Setting} 
\label{sec:experimental_setting}
To evaluate the proposed OATM, we design extensive experiments to tune the hyper-parameters of two most widely used deep learning structures, i.e., the Recurrent Neural Networks (RNNs) and the Convolutional Neural Networks (CNNs). Both of the two deep learning structures are employed on three real-world applications: 1) a human intention recognition task based on the Electroencephalography (EEG) signals; 2) activity recognition based on wearable sensors like Inertial Measurement Unit (IMU); 3) activity recognition based on pervasive sensors like Radio Frequency IDentification (RFID).

\subsection{Data Setting} 
\label{sub:data_setting}
The proposed OATM is evaluated over three different tasks on three benchmark datasets. Each dataset is divided into a training set (80\%) and a testing set (20\%).

\subsubsection{EEG-based Intention Recognition.} 
We select the widely used EEG dataset from PhysioNet eegmmidb database\footnote{\url{https://www.physionet.org/pn4/EEGmmidb/}} which contains 5 different categories. In this paper, we choose a subset of eegmmidb which contains 28,000 EEG samples. Every sample is a vector with 64 elements corresponding to 64 channels.

\subsubsection{IMU-based Activity Recognition.} This dataset is collected by 9 participants \cite{fida2015real}, which contains 1200000 samples. 8 ADLs are selected as a subset of our paper. The activity is measured by 3 IMUs and each IMU collects sensor signal with 14 dimensions including two 3-axis accelerometers, one 3-axis gyroscopes, one 3-axis magnetometers, and one thermometer.

\subsubsection{RFID-based Activity Recognition.} We collect the signals from passive RFID tags \cite{yao2017compressive} and have 3100 samples in total. 21 activities, including 18 ADLs (Activity of Daily Living) and 3 abnormal falls, are performed by 6 subjects. Each sample has 12 dimensions corresponding to 12 RFID tags. RSSI measures the power present in a received radio signal, which is a convenient environmental measurement in ubiquitous computing.

\subsection{Deep Learning Structures} 
\label{sub:deep_learning_structures}
In this section, we briefly describe RNN and CNN structures and then introduce the key hyper-parameters that will be tuned in the experiments.

\subsubsection{RNN Structure} 
\label{sub:rnn}
The RNN \cite{li2018recurrent}, one of the most widely-used deep neural networks, is generally employed to explore the feature dependencies over time dimension through an internal state of the network.
Unlike feed-forward neural networks, RNNs can use their internal memory to process arbitrary sequences of inputs and exhibit dynamic temporal behavior. Such characteristic ensures RNNs achieve excellent performance in time-series tasks such as speech recognition and natural language processing.

\begin{figure}[t!]
\includegraphics[width=\linewidth]{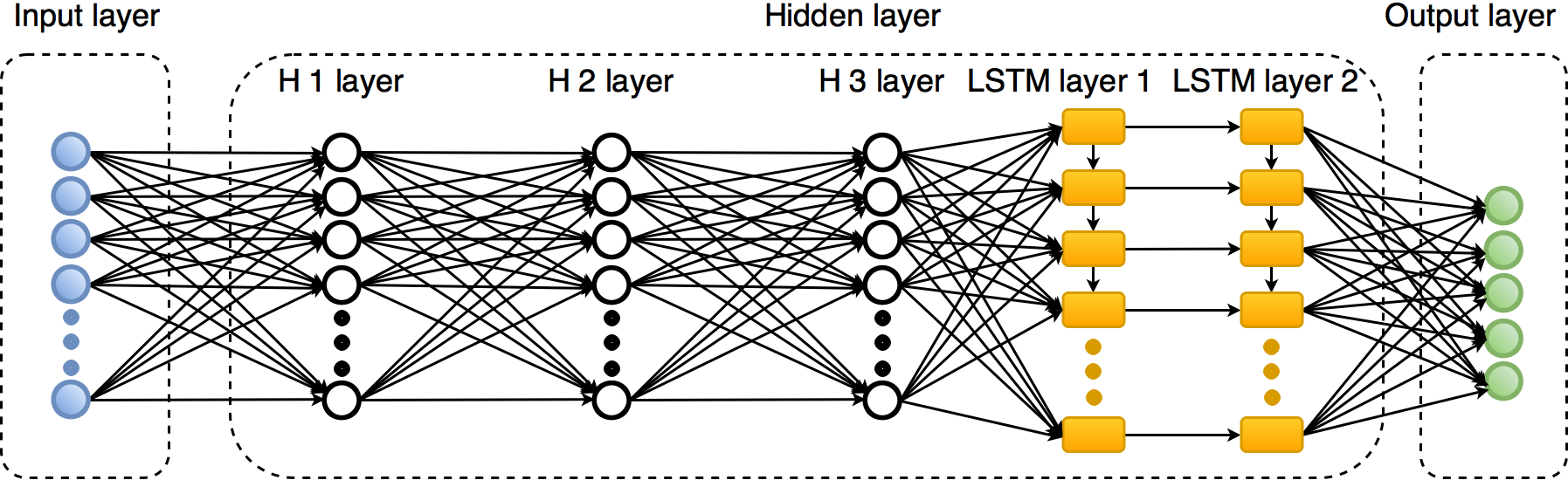}
\caption{The schematic diagram of RNN structure. `H' denotes Hidden, where, for example, the \textit{H 1 layer} denotes the first hidden layer. 
} 
\label{fig:rnn}
\vspace{-3mm}
\end{figure}

The RNN structure used in this paper is shown in Figure~\ref{fig:rnn}.
In the hidden layer, to implement the recurrent function, two LSTM (Long Short-Term Memory) layer is concentrated. LSTM is a simple cell structure which can be used to build a recurrent neural network. Different from other fully connected layers, LSTM layer is composed of cells (shown as rectangles) instead of neural nodes (shown as circles).

In this RNN structure,
based on the deep learning hyper-parameters tuning experience, the learning rate, the regularization, and the number of nodes in each hidden layer are key factors affecting the algorithm performance. The loss is calculated by cross-entropy function, and the regularization method is $\ell_2$ norm with the coefficient $\lambda$, The loss is finally optimized by the AdamOptimizer algorithm. In summary, we choose four factors as to-be-tuned hyper-parameters: the learning rate $lr$, the regularization coefficient $\lambda$, the number of hidden layers $n_l$, and the number of nodes\footnote{Assume all the hidden layers have the same fixed number of nodes.} in each hidden layer $n_n$.

\subsubsection{CNN Structure} 
\label{sub:cnn}
\begin{figure}[t!]
\centering
\includegraphics[width=\linewidth]{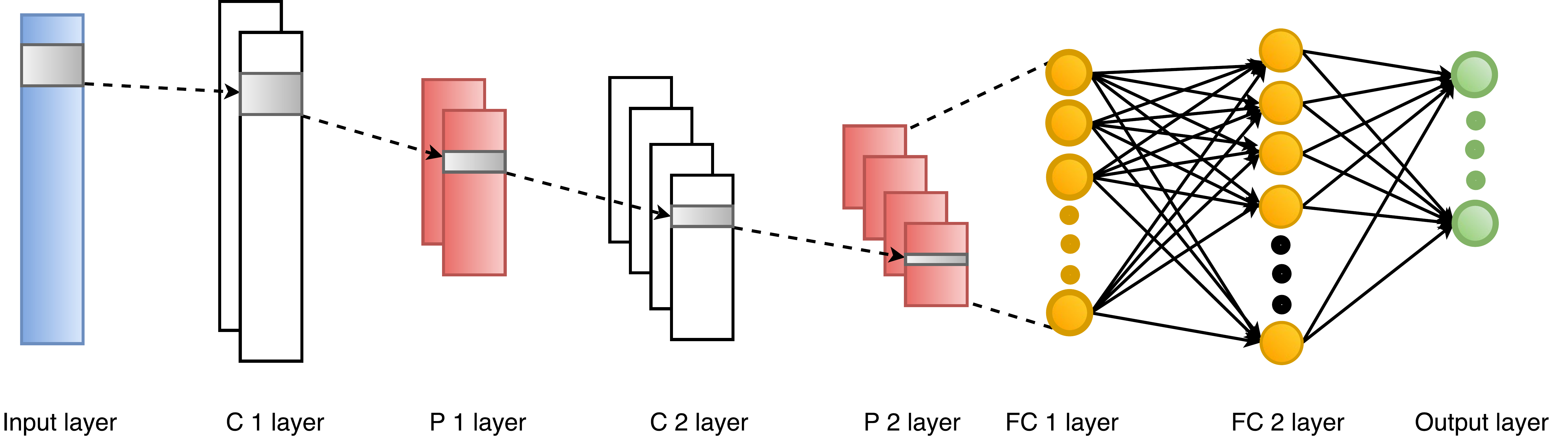}
\caption{The schematic diagram of CNN structure. 
\textit{C}, \textit{P}, and \textit{FC} denote convolutional layer, pooling layer, and fully connected layer, respectively.
} 
\label{fig:cnn}
\vspace{-5mm}
\end{figure}

The CNN is another popular deep neural network, which shows strong ability to capture the latent spatial relevance of the input data and has been demonstrated in a wide range of research topics such as computer vision \cite{gu2018recent}. The CNN structure contains three categories of components: the convolutional layer, the pooling layer, and the fully connected layer. Each component may appear one or multiple times in a CNN. 
As shown in Figure~\ref{fig:cnn}, the schematic diagram of CNN is stacked in the following order: the input layer, the first convolutional layer, the first pooling layer, the second convolutional layer, the second pooling layer, the first fully connected layer, the second fully connected layer, and the output layer. The loss function, regularization method, and optimizer are the same as those in the RNN structure. Based on hyper-parameters tuning experience on CNN, we choose four most crucial factors to be tuned by OATM: the learning rate $lr'$, the filter size $f'$, the number of convolutional and pooling layers $n_l'$\footnote{We consider each convolutional layer and the following pooling layer as whole.}, and the number of nodes $n_n'$ in the second fully connected layer.

\section{Results and Analysis} 
\label{sec:results_analysis} 
In this section, we present the hyper-parameter tuning results by OATM and compare to the state-of-the-art methods over a very comprehensive scenario which contains two deep learning structures working on three datasets. For simplification, we set the same hyper-parameter ranges for the three datasets. \textit{All the codes are open-sourced, please check the code for the training details which are not presented here due to page limitation, }

\subsection{Overall Comparison} 
\label{sub:overall_comparison}
In this section, we compare the proposed OATM with the most competitive state-of-the-art hyper-parameter tuning approaches including two manually methods (grid search and random search) and an automated one (Bayesian Optimization). It's easy to compute that there are $81=3^4$ exhausted combinations in grid search since we have four factors with three levels of the hyper-parameters. Thus, grid search requires 81 runnings to get the optimal hyper-parameters. On the other hand, our method requires only 9 runnings described in the corresponding orthogonal array table (detailed in Section~\ref{sub:hyper_parameters_tuning_in_rnn}). Due to the numbers of runnings in random search and Bayesian Optimization are manually set, they are set as 9 runnings which is same with our method in order to keep fair comparison. The baselines are introduced here:

\begin{itemize}
    \item Grid search simply goes through all the possible combinations according to the values provided which is exhaustive \cite{NIPS2011_4443}.
    \item  Random search randomly picks combinations from all possible ones. It may not find a decent combination but is widely adopted in industry for the high-efficiency \cite{Andradottir2015}.
    \item Bayesian optimization uses a Gaussian process to minimize the loss function in order to maximize performance \cite{NIPS2012_4522}.
\end{itemize}

\begin{table}[t!]
\caption{Comparison with the state-of-the-art methods over three datasets and two deep learning architectures. The F1 $\sim$ F4 represent four tuning factors. Acc, Prec and F-1 denote accuracy, precision and F-1 score, respectively. }
\label{tab:results_comparison}
\resizebox{\linewidth}{!}{
\begin{tabular}{lll|cccc|cllcll}
\hline
\multirow{2}{*}{\textbf{Data}} & \multirow{2}{*}{\textbf{Models}} & \multirow{2}{*}{\textbf{Methods}} & \multicolumn{4}{c|}{\textbf{\begin{tabular}[c]{@{}c@{}}Optimal Factors\end{tabular}}} & \multicolumn{6}{c}{\textbf{Metrics}} \\ \cline{4-13}
 &  &  & \textbf{F1} & \textbf{F2} & \textbf{F3} & \textbf{F4} & \textbf{\#-Runnings} & \textbf{Time (s)} & \textbf{Acc} & \textbf{Prec} & \textbf{Recall} & \textbf{F-1} \\ \hline
 \multirow{8}{*}{\textbf{EEG}} & \multirow{4}{*}{\textbf{RNN}} & \textbf{Grid} & 0.005 & 0.004 & 6 & 64 & 81 & 6853.6 & 0.9251 & 0.9324 & 0.9139 & 0.9231 \\
 &  & \textbf{Random} & 0.01 & 0.008 & 6 & 32 & 9 & 766.8 & 0.7941 & 0.8003 & 0.7941 & 0.7947 \\
 &  & \textbf{BO} & 0.0135 & 0.0049 & 5 & 32 & 9 & 703.4 & 0.718 & 0.7246 & 0.6474 & 0.6838 \\
 &  & \textbf{Ours} & 0.005 & 0.004 & 6 & 64 & 9 & 821.9 & 0.925 & 0.9335 & 0.9223 & 0.9279 \\ \cline{2-13}
 & \multirow{4}{*}{\textbf{CNN}} & \textbf{Grid} & 0.005 & 4 & 3 & 192 & 81 & 31891.5 & 0.828 & 0.8137 & 0.8256 & 0.8269 \\
 &  & \textbf{Random} & 0.003 & 2 & 1 & 128 & 9 & 662.8 & 0.7268 & 0.7277 & 0.7269 & 0.7266 \\
 &  & \textbf{BO} & 0.001 & 4 & 3 & 139 & 9 & 721.9 & 0.7244 & 0.7302 & 0.7244 & 0.7263 \\
 &  & \textbf{Ours} & 0.003 & 4 & 1 & 128 & 9 & 680.4 & 0.797 & 0.7969 & 0.8112 & 0.8003 \\ \hline
\multirow{8}{*}{\textbf{IMU}} & \multirow{4}{*}{\textbf{RNN}} & \textbf{Grid} & 0.005 & 0.004 & 6 & 96 & 81 & 3027.2 & 0.9936 & 0.9909 & 0.9976 & 0.9971 \\
 &  & \textbf{Random} & 0.015 & 0.004 & 4 & 32 & 9 & 1008.5 & 0.9139 & 0.9209 & 0.9412 & 0.9156 \\
 &  & \textbf{BO} & 0.0132 & 0.0041 & 4 & 48 & 9 & 1078.8 & 0.9872 & 0.9877 & 0.9851 & 0.9863 \\
 &  & \textbf{Ours} & 0.005 & 0.004 & 6 & 64 & 9 & 1138.2 & 0.9913 & 0.9924 & 0.9905 & 0.9919 \\ \cline{2-13}
 & \multirow{4}{*}{\textbf{CNN}} & \textbf{Grid} & 0.003 & 2 & 1 & 128 & 81 & 41804.9 & 0.9732 & 0.9708 & 0.9708 & 0.9707 \\
 &  & \textbf{Random} & 0.003 & 2 & 2 & 128 & 9 & 7089.2 & 0.9692 & 0.9691 & 0.9692 & 0.9691 \\
 &  & \textbf{BO} & 0.0012 & 2 & 2 & 192 & 9 & 6559.7 & 0.9696 & 0.9702 & 0.9701 & 0.9701 \\
 &  & \textbf{Ours} & 0.003 & 2 & 2 & 128 & 9 & 6809.8 & 0.9702 & 0.9699 & 0.9703 & 0.9702 \\ \hline
\multirow{8}{*}{\textbf{RFID}} & \multirow{4}{*}{\textbf{RNN}} & \textbf{Grid} & 0.005 & 0.008 & 6 & 96 & 81 & 2846.1 & 0.9342 & 0.9388 & 0.9201 & 0.9252 \\
 &  & \textbf{Random} & 0.005 & 0.012 & 4 & 32 & 9 & 642.3 & 0.8891 & 0.9138 & 0.8826 & 0.8895 \\
 &  & \textbf{BO} & 0.0142 & 0.0093 & 6 & 79 & 9 & 452.2 & 0.9071 & 0.8556 & 0.8486 & 0.8436 \\
 &  & \textbf{Ours} & 0.005 & 0.008 & 6 & 64 & 9 & 497.1 & 0.9134 & 0.9138 & 0.9029 & 0.9162 \\ \cline{2-13}
 & \multirow{4}{*}{\textbf{CNN}} & \textbf{Grid} & 0.005 & 4 & 2 & 192 & 81 & 7890.8 & 0.9316 & 0.9513 & 0.9316 & 0.9375 \\
 &  & \textbf{Random} & 0.005 & 2 & 1 & 128 & 9 & 1210.3 & 0.8683 & 0.9113 & 0.8684 & 0.8779 \\
 &  & \textbf{BO} & 0.005 & 5 & 3 & 64 & 9 & 872.9 & 0.9168 & 0.9058 & 0.9194 & 0.9086 \\
 &  & \textbf{Ours} & 0.005 & 4 & 3 & 192 & 9 & 980.3 & 0.9235 & 0.9316 & 0.9188 & 0.9326\\ \hline
\end{tabular}
}
\vspace{-3mm}
\end{table}

The hyper-parameter levels are selected based on empirical values. For grid search, random search, and our OATM, the empirical values are discrete as listed in Table~\ref{tab:fl_rnn} (take eegmmidb as an example). For Bayesian Optimization, the hyper-parameter ranges from the maximum and minimum of each factor. For instance, the $lr$ ranges from $[0.005, 0.015]$. All the experiments are implemented in NVIDIA Titian X (Pascal) GPU and each reported value is the average of five runnings under the same setting.

The comparison results are shown in Table~\ref{tab:results_comparison}. It can be observed that: 
\begin{itemize}
    \item under the same running numbers (9 runnings), our method outperforms the random search and Bayesian Optimization over all the datasets and deep learning architectures;
    \item our method performs slightly lower than grid search but still competitive, however, take EEG dataset with RNN as an example, our approach saves 88\% tuning time which is indicated from that the OATM only requires 9 runnings and costs 821.9s while grid search requires 81 runnings and 6853.6s;
    \item the optimal factors selected by our method approximate to the global optimal factors selected by grid search. 
\end{itemize}

\subsection{Case Study in RNN and CNN} 
\label{sub:hyper_parameters_tuning_in_rnn}
In this section, we take EEG classification as an example to present the detailed procedure of OATM in RNN and CNN architecture. The overall paradigm can be divided into five steps.

\vspace{-5mm}
\subsubsection{Step 1: Build the F-L table} 
\label{sub:setp_1}
According to the description in Section~\ref{sub:rnn}, OATM will work on four different hyper-parameters (factors): the learning rate $lr$, the $l$-2 norm coefficient $\lambda$, the number of hidden layers $n_l$, and the number of nodes $n_n$.
 The number of levels $h$ is set to be 3 which could be much larger in real-world applications.
 Based on the related work and tuning experience \cite{zhang2017intent}, the empirical values are shown in Table~\ref{tab:fl_rnn}.

\begin{table}[t]
\centering
\caption{Factor-Level table of RNN and CNN.
}
\label{tab:fl_rnn}
\begin{tabular}{cccccc}
\hline
\multirow{4}{*}{\textbf{RNN}} &  & \textbf{Factor 1 ($lr$) } & \textbf{Factor 2 ($\lambda$) } & \textbf{Factor 3 ($n_l$) } & \textbf{Factor 4 ($n_n$) } \\ \cline{2-6}
& \textbf{Level 1} & 0.005 & 0.004 & 4 & 32 \\
& \textbf{Level 2} & 0.01 & 0.008 & 5 & 64 \\
& \textbf{Level 3} & 0.015 & 0.012 & 6 & 96 \\ \hline
\multirow{4}{*}{\textbf{CNN}} &  & \textbf{Factor 1 ($lr'$) } & \textbf{Factor 2 ($f'$) } & \textbf{Factor 3 ($n_l'$) } & \textbf{Factor 4 ($n_n'$) } \\ \cline{2-6}
& \textbf{Level 1} & 0.001 & [1,2] & 1 & 64 \\
& \textbf{Level 2} & 0.003 & [1,4] & 2 & 128 \\
& \textbf{Level 3} & 0.005 & [1,6] & 3 & 192  \\ \hline
\end{tabular}
\vspace{-3mm}
\end{table}

\vspace{-5mm}
\subsubsection{Step 2: OATM table} 
\label{sub:step_2_oat_table}
Then, choosing a suitable Orthogonal Array table which contains 4 factors and 3 levels for our experiments in this link\footnote{\url{https://www.york.ac.uk/depts/maths/tables/taguchi_table.htm}} wich contains 9 combinations.
The OATM table should satisfy two basic principles: i) in each column, different levels have the same appear times; ii) in any two randomly-selected columns, nine differently-ordered element combinations
are completed and balanced. 

\vspace{-5mm}
\subsubsection{Step 3: Run the experiments} 
\label{sub:step_3_run_the_experiments}
 Following the OATM table, run the 9 experiments and record the classification accuracy.
In our case, each experiment runs 5 times with the corresponding average accuracy recorded. Each experiment is trained for 1,000 iterations to guarantee the convergence.

\vspace{-5mm}
\subsubsection{Step 4: Range analysis} 
\label{sub:step_4_range_analysis}
This is the key step of Orthogonal Array Tuning. The overall range analysis procedure and results are shown in Table~\ref{tab:ra_rnn}. The first 9 rows are measured and recorded in Step 3. 
$R_{leveli}$ denotes the sum of accuracy under level $i$. For example, $R_{level1}$ in factor 1 is the sum of the accuracy in the first 3 rows ($0.196=0.875+0.8+0.521$), where factor 1 is on level 1. $A_{leveli}$ denotes the average accuracy of level $i$, calculated by $A_{leveli} =R_{leveli}/h$. In the above example, we calculate $A_{level1}$ as $0.732=2.196/3$. Lowest and highest accuracy values, measuring the maximum and minimum of $A_{leveli}$ respectively, are used to calculate the {\it range} of $A_{leveli}$. The importance denotes how important the factor is, which is ranked by the range value. 
{\it Best level} is the selected optimal level based on the {\it Highest Acc} while {\it Optimal Value} represents the corresponding value of the best level.

\begin{table}[t!]
\centering
\caption{Range analysis of RNN
}
\label{tab:ra_rnn}
\resizebox{0.9\linewidth}{!}{
\begin{tabular}{c|ccccc}
\hline
\textbf{Row No.} & \textbf{Factor 1 ($lr$) } & \textbf{Factor 2 ($\lambda$) } & \textbf{Factor 3 ($n_l$) } & \textbf{Factor 4 ($n_n$) } & \textbf{Acc} \\ \hline 
\textbf{1} & 0.005 & 0.004 & 4 & 32 & 0.875 \\ 
\textbf{2} & 0.005 & 0.008 & 5 & 64 & 0.8 \\ 
\textbf{3} & 0.005 & 0.012 & 6 & 96 & 0.521 \\ 
\textbf{4} & 0.01 & 0.004 & 5 & 96 & 0.888 \\ 
\textbf{5} & 0.01 & 0.008 & 6 & 32 & 0.797 \\ 
\textbf{6} & 0.01 & 0.012 & 4 & 64 & 0.451 \\ 
\textbf{7} & 0.015 & 0.004 & 6 & 64 & {\bf 0.897} \\ 
\textbf{8} & 0.015 & 0.008 & 4 & 96 & 0.335 \\ 
\textbf{9} & 0.015 & 0.012 & 5 & 32 & 0.471 \\ \hline
\textbf{$R_{level1}$} & 2.196 & 2.66 & 1.661 & 2.143 &  \\
\textbf{$R_{level2}$} & 2.136 & 1.932 & 2.159 & 2.148 &  \\
\textbf{$R_{level3}$} & 1.703 & 1.443 & 2.215 & 1.744 &  \\ \hline
\textbf{$A_{level1}$} & \textbf{0.732} & \textbf{0.887} & 0.554 & 0.714 &  \\
\textbf{$A_{level2}$} & 0.712 & 0.644 & 0.720 & \textbf{0.716} &  \\
\textbf{$A_{level3}$} & 0.568 & 0.481 & \textbf{0.738} & 0.581 &  \\ \hline
\textbf{Lowest Acc} & 0.568 & 0.481 & 0.554 & 0.581 &  \\
\textbf{Highest Acc} & 0.732 & 0.887 & 0.738 & 0.716 &  \\
\textbf{Range} & 0.164 & 0.406 & 0.184 & 0.135 &  \\
\textbf{Importance} & \multicolumn{4}{c}{\textbf{$lambda>n_l>lr>n_n$}} &  \\ \hline
\textbf{Best Level} & \textbf{Level 1} & \textbf{Level 1} & \textbf{Level 3} & \textbf{Level 2} & \textbf{} \\
\textbf{Optimal Value} & \textbf{0.005} & \textbf{0.004} & \textbf{6} & \textbf{64} & \textbf{0.925} \\ \hline
\end{tabular}
}
\vspace{-3mm}
\end{table}

\vspace{-5mm}
\subsubsection{Step 5: Run the optimal setting} 
\label{sub:step_5_run_the_optimal_setting}
Since the best level is given by the range analysis in the previous step, we run the experiment with the optimal hyper-parameters ($lr=0.004$, $\lambda=0.005$, $n_l=6$, and $n_n=64$) and finally we got the optimal accuracy as $0.925$. We can observe that:
\begin{itemize}
    \item The optimal accuracy $0.925$ is higher than the maximum of the accuracy ($0.897$) in the OATM experiments, which demonstrates that the OATM is enabled to approximate the global optimal instead of the local optimal.
    \item The importance of each factor is ranked through the range analysis: $lambda>n_l>lr>n_n$, which can guide the researcher to grasp the dominating variable in the RNN structure and be helpful in the future development.
\end{itemize}

The OATM paradigm of CNN is similar to RNN. Here, we only report the F-L table (Table~\ref{tab:fl_rnn}) and the range analysis table (Table~\ref{tab:ra_cnn}).

\begin{table}[t]
\centering
\caption{Range analysis of CNN
}
\label{tab:ra_cnn}
\resizebox{0.9\linewidth}{!}{
\begin{tabular}{c|ccccc}
\hline
\textbf{Row No.} & \textbf{Factor 1 ($lr'$)} & \textbf{Factor 2 ($f'$)} & \textbf{Factor 3 ($n_l'$)} & \textbf{Factor 4 ($n_n'$)} & \textbf{Acc} \\ \hline 
\textbf{1} & 0.001 & [1,2] & 1 & 64 & 0.707 \\ 
\textbf{2} & 0.001 & [1,4] & 2 & 128 & 0.771 \\ 
\textbf{3} & 0.001 & [1,6] & 3 & 192 & 0.775 \\ 
\textbf{4} & 0.003 & [1,2] & 2 & 192 & 0.779 \\ 
\textbf{5} & 0.003 & [1,4] & 3 & 64 & 0.752 \\ 
\textbf{6} & 0.003 & [1,6] & 1 & 128 & \textbf{0.797} \\ 
\textbf{7} & 0.005 & [1,2] & 3 & 128 & 0.784 \\ 
\textbf{8} & 0.005 & [1,4] & 1 & 192 & 0.782 \\ 
\textbf{9} & 0.005 & [1,6] & 2 & 64 & 0.756 \\ \hline
\textbf{$R_{level1}$} & 2.253 & 2.27 & 2.993 & 2.215 &  \\
\textbf{$R_{level2}$} & 2.328 & 2.305 & 2.306 & 2.352 &  \\
\textbf{$R_{level3}$} & 2.322 & 2.328 & 2.311 & 2.336 &  \\ \hline
\textbf{$A_{level1}$} & 0.751 & 0.757 & \textbf{0.998} & 0.738 &  \\
\textbf{$A_{level2}$} & \textbf{0.776} & 0.768 & 0.769 & \textbf{0.784} &  \\
\textbf{$A_{level3}$} & 0.774 & \textbf{0.776} & 0.770 & 0.779 &  \\ \hline
\textbf{Lowest Acc} & 0.751 & 0.757 & 0.769 & 0.738 &  \\
\textbf{Highest Acc} & 0.776 & 0.776 & 0.998 & 0.784 &  \\
\textbf{Range} & 0.025 & 0.019 & 0.229 & 0.046 & \cellcolor[HTML]{FFFFFF}\textbf{} \\
\textbf{Importance} & \multicolumn{4}{c}{$n_l'>n_n'>lr'>f'$} &  \\ \hline
\textbf{Best Level} & \cellcolor[HTML]{FFFFFF}\textbf{Level 2} & \cellcolor[HTML]{FFFFFF}\textbf{Level 3} & \cellcolor[HTML]{FFFFFF}\textbf{Level 1} & \cellcolor[HTML]{FFFFFF}\textbf{Level 2} &  \\
\textbf{Optimal Value} & \textbf{0.003} & \textbf{[1,6]} & \textbf{1} & \textbf{128} & \textbf{0.797} \\ \hline
\end{tabular}
}
\vspace{-3mm}
\end{table}

\section{Discussion and Conclusion} 
\label{sec:conclusion}

In this paper, we present an efficient OATM approach for hyper-parameter tuning in the context of deep learning. 
The proposed OATM is evaluated over two popular deep learning structures(RNN and CNN) over three real-world datasets. The experiment results show that our approach outperforms state-of-the-art hyper-parameter tuning methods.

One disadvantage of OATM is that it requires the empirical values as prerequisites. The values of the F-L table should be chosen appropriately. However, this is the common drawback of all the tuning methods. For instance, the hyper-parameter ranges in Bayesian Optimization are also pre-defined based on empirical values.


\bibliographystyle{splncs03}
\bibliography{OA.bib}


\end{document}